\documentclass{article}

\usepackage[preprint]{neurips_2024}

\usepackage{enumitem}

\usepackage[utf8]{inputenc} 
\usepackage[T1]{fontenc} 
\usepackage{hyperref} 
\usepackage{url} 
\usepackage{booktabs} 
\usepackage{amsfonts} 
\usepackage{nicefrac} 
\usepackage{microtype} 
\usepackage{xcolor} 
\usepackage{longtable}
\usepackage{natbib}
\usepackage{graphicx}
\usepackage{amsmath}
\usepackage{amssymb}
\usepackage{mathtools}
\usepackage{amsfonts}

\hypersetup{
 colorlinks  = true,
 urlcolor   = blue,
 linkcolor  = blue,
 citecolor  = blue
}

\title{The Mysterious Case of Neuron 1512: Injectable Realignment Architectures Reveal Internal Characteristics of Meta's Llama 2 Model}

\author{
 Brenden Smith, ~ Dallin Baker, ~ Clayton Chase, \\ ~ Myles Barney, ~ Kaden Parker, ~ Makenna Allred, ~ Peter Hu, ~ Alex Evans, ~ Nancy Fulda \\
 \bf{Brigham Young University}
}

\begin{document}
\maketitle

\begin{abstract}

Large Language Models (LLMs) have an unrivaled and invaluable ability to "align" their output to a diverse range of human preferences, by mirroring them in the text they generate. The internal characteristics of such models, however, remain largely opaque. This work presents the Injectable Realignment Model (IRM) as a novel approach to language model interpretability and explainability. Inspired by earlier work on Neural Programming Interfaces, we construct and train a small network -- the IRM -- to induce emotion-based alignments within a 7B parameter LLM architecture. The IRM outputs are injected via layerwise addition at various points during the LLM's forward pass, thus modulating its behavior without changing the weights of the original model. This isolates the alignment behavior from the complex mechanisms of the transformer model. Analysis of the trained IRM's outputs reveals a curious pattern. Across more than 24 training runs and multiple alignment datasets, patterns of IRM activations align themselves in striations associated with a neuron's index within each transformer layer, rather than being associated with the layers themselves. Further, a single neuron index (1512) is strongly correlated with all tested alignments. This result, although initially counterintuitive, is directly attributable to design choices present within almost all commercially available transformer architectures, and highlights a potential weak point in Meta's pretrained Llama 2 models. It also demonstrates the value of the IRM architecture for language model analysis and interpretability.

 \vspace{5mm}

\end{abstract}

\section{Introduction}
\label{intro}

Due to their size and high level of abstraction, language models often function as "black boxes" that offer little insight into their internal workings. This limits the ability of researchers and practitioners to draw connections between a model's behavior and the specific content of its conditioning prompt or the parametric knowledge stored in its weight matrices. It also prevents researchers from understanding how the activations of specific neuron groups relate to real-world concepts such as ethnicity, emotions, mathematical relations, or physical objects. This opacity limits utility of language models for scientific purposes and restricts their deployment in contexts that preclude decision-making based on legally protected categories.

Understanding why models produce certain outputs is arduous and complex. Early approaches to language model analysis involved simplification techniques like dimension reduction \citep{chipman2005interpretable}, but these techniques typically fall short when applied to larger and complex models -- only serving as approximations and losing the fidelity of the original model. More recently \cite{towardsmonosemanticity} presented a dictionary learning approach that successfully identified polysemantic features associated with specific real-world topics, but the resulting features are themselves difficult to interpret, and the process relies on meta-analysis via a language model much larger than the one being studied. This work presents an alternative that allows targeted inspection of LLM activation patterns using a construct much smaller than the model being analyzed.

We present the Injectable Realignment Model (IRM): a small, fully-connected neural network used to manipulate the \textit{activations}, or hidden layer outputs, of a much larger language model. During each forward pass of the language model, the IRM receives as input the activation values of the language model's initial attention layer and produces outputs that are summed with the outputs of the language model's hidden layers during inference. The IRM thus manipulates the language model's behavior without changing the parameters of the model itself. We intuit that the IRM's outputs carry valuable information about the larger model's parametric structure, since the successful \textit{induction} of an aligned behavior relies on the detection (and manipulation) of neurons \textit{associated} with that behavior.

We apply our IRM method to a 7B-parameter Llama 2 chat model (\cite{llama-2-7B-chat}) and show that 2.1B-parameter IRM is able to induce text generations with characteristics related to two well-studied emotions: anger and sadness (Section \ref{sec_isolated_alignment}). We then analyze the pattern of IRM outputs (Section \ref{sec:activation_patterns}) to determine which of Llama's neural activations are most strongly associated with the induced behaviors. Surprisingly, we discover that a single neuron index is strongly correlated with both anger and sadness across multiple data runs and across nearly all transformer blocks within the Llama-2 model. This result is both counterintuitive and strongly reminiscent of \cite{radford2017learning}'s observation of a "sentiment neuron". We discuss this odd behavior, and show why it follows naturally from an analysis of the commonly-used transformer architecture, in Sections \ref{4.3} and \ref{sec:1512}. Taken together, the contributions of this paper are as follows:

\begin{itemize}[itemsep=0.1pt]
  \item \textbf{Isolated Alignment.} We present the Injectable Realignment Model, a neural architecture with potential to separate a desired alignment strategy from the model itself. This isolation of a desired bias is a step toward model-agnostic alignment programs.
  \item \textbf{Interpretability via Alignment Injection.} We leverage a trained IRM, which has approximately 1/3 as many parameters as the model it influences, as a tool to study architectural features of the Llama-2-chat model, and discover a particular neuron with disproportionate influence over inducted alignments.
  \item \textbf{Vertical Continuity:} Analysis via Injectable Realignment Models shows that the Llama-2 language model exhibits vertical continuity, a phenomenon in which aligned behaviors are associated with the same neuron index across multiple transformer blocks. This phenomenon is induced by skip connections (also called residual connections) within the Llama-2 architecture, and is likely to exist in all transformer-based language models.
  \item \textbf{Overburdened Language Modeling:} Informed by the observation of vertical continuity, we draw parallels between the Llama-2 language modeling head and known weaknesses of variational autoencoders, in which the decoder portion of the model carries an unequal share of the model's generalization capacity. These observations are paired with suggestions for stronger language modeling capabilities within transformer architectures.
\end{itemize}

Although the IRM architecture shows potential for fluent and sophisticated alignment induction within language models with 70B+ parameters, the initial probing experiments reported here explore only simplified alignment tasks, and only in models with 7B parameters. The application of Injectable Realignment Models in more complex scenarios, and the assurance of full coherence in the aligned model, is left to future work.

\section{Related work}
\label{rel_work}

The quest to improve alignment and interpretability in Large Language Models (LLMs) is a crucial component of developing trustworthy, explainable, and ethical artificial intelligence systems. Recent breakthroughs in LLM development, fueled by transformer architectures (\citet{NIPS2017_3f5ee243}), have revolutionized natural language processing, demonstrating remarkable abilities in language generation, translation, and complex task execution (\citet{Liu_2023}). Models, like those introduced by \citet{touvron2023Llama}, showcased the fine-tuned Llama chat model and new fine-tuning approaches, and the work on scaling model size done by \citet{chung2022scaling}, showcased the capacity of LLMs to process extensive data and tackle complex tasks. These models, however, are prone to hallucinations (generating factually incorrect text) and biased outputs, necessitating methods to guide them towards more reliable and ethical responses.

Prompt engineering, as investigated by \citet{journals/corr/abs-2005-14165} and \citet{DBLP:journals/corr/abs-2102-09690}, alongside fine-tuning techniques from \citet{dathathri2020plug} and \citet{DBLP:journals/corr/abs-2101-00190}, are crucial to elicit preferred responses in an LLM. Prompt engineering allows a researcher to shape the model's paradigm, protecting users from harmful content and increasing the model's capability to generate accurate responses. Despite the impressive capabilities of models like Llama (\citet{touvron2023Llama}) and GPT-4 (\citet{openai2024gpt4}), their interpretability remains inaccessible. Efforts to enhance understanding of LLMs, including studies by \citet{xie-etal-2023-proto}, which fine-tuned models to be more easily understood, and \citet{voita-etal-2019-bottom}, which analyzed information flow within transformer layers, aim to dissect the transformer architecture and its layer-specific functionalities.

Furthering the discourse on model interpretability, \citet{tigges2023linear} explores sentiment representation within LLMs, while \citet{conmy2023automated} introduces automated methods for identifying neural network "circuits." Critical analyses by \citet{wen2023transformers} and in-depth examinations of model efficacy and explainability by \citet{10356671} and \citet{bills2023language} contribute to the evolving understanding of LLM behavior and interpretation by laying the groundwork for techniques aimed at increasing model transparency and understanding.

The limitations of current interpretability methods are scrutinized by \citet{friedman2023interpretability} which showed that widely accepted methods of interpreting model behavior -- especially simplified model representations -- were not capable of "accurately captur[ing] the model's behavior out of distribution", emphasizing the need for precise representation of model behaviors. Studies like those by \citet{gurnee2023finding} and \citet{mcdougall2023copy} delve into the neural representation and architectural nuances influencing LLM outputs,discovering that different layers within the transformer architecture may serve specific, discrete functions. Additionally, \citet{liu2023devil} and \citet{hase2024does} investigate bias mitigation and knowledge editing within LLMs, enhancing the understanding of model manipulation.

Our paper introduces a novel mechanism aimed at improving LLM alignment and interpretability similar to approaches by \citet{li2023emergent} and \citet{vig2020causal} in that we focus on fine-tuning our models to achieve a specific goal, as done by \citet{li2023emergent}, and analyze the model's structure in relation to its behavior, as introduced by \citet{vig2020causal}. However, our research differs in our utilization of the IRM to directly modify the LLM during the training and inference process. By integrating a small neural network into the transformer structure, the IRM enables targeted output modifications, facilitating a granular analysis of how different layers affect language generation. This approach not only aligns LLM behavior with desired standards but also enriches our comprehension of the model’s internal dynamics, contributing to a more nuanced understanding of LLM architecture and functionality.

\section{Methods}
\label{methods}

Commercial language models are predominately based on the transformer architecture introduced by \cite{NIPS2017_3f5ee243}. Our work leverages Meta's Llama-2 7B language model, which contains 32 transformer blocks, each comprised of a self-attention module paired with a linear feed-forward network using a SwiGLU activation function \citep{shazeer2020glu}. In this paper, we concern ourselves with the floating point values output by the self-attention module, and term them the \textit{activations} of that block or layer of the model; and we metaphorically refer to these activation regions as \textit{neurons}. Such metaphorical terminology harks to early work in multi-layer perceptions \cite{rosenblatt1962principles}, and is often useful in conceptualizing the inner workings of artificial neural networks.

Our Injectable Realignment Model, originally inspired by \citep{brown2020towards}'s work on Neural Programming Interfaces, influences a pre-trained Llama-2 model by perturbing its neuron activations during inference. However, whereas Neural Programming Interfaces are trained using a generative-adversarial framework \citep{goodfellow2014generative} and rely on input activations drawn from many different transformer blocks within the host model, Injectable Realignment Models use a straightforward fine-tuning methodology and rely on only a single set of activations as input. Their primary purpose is language model interpretability rather than controlled text generation. 

On each forward pass, the IRM receives as inputs the activations of the Llama model's initial attention layer. The IRM then produces values that are summed with the post-attention neurons of each transformer block in the Llama model. (Optionally, the IRM outputs can be injected into only a subset of the transformer blocks.) During training, the Llama model's weights are frozen and the desired model alignment is induced solely via the IRM's learned permutations.

\subsection{Injected Model Structure}

\begin{figure}[htbp]
 \centering
 \includegraphics[width=0.6\textwidth]{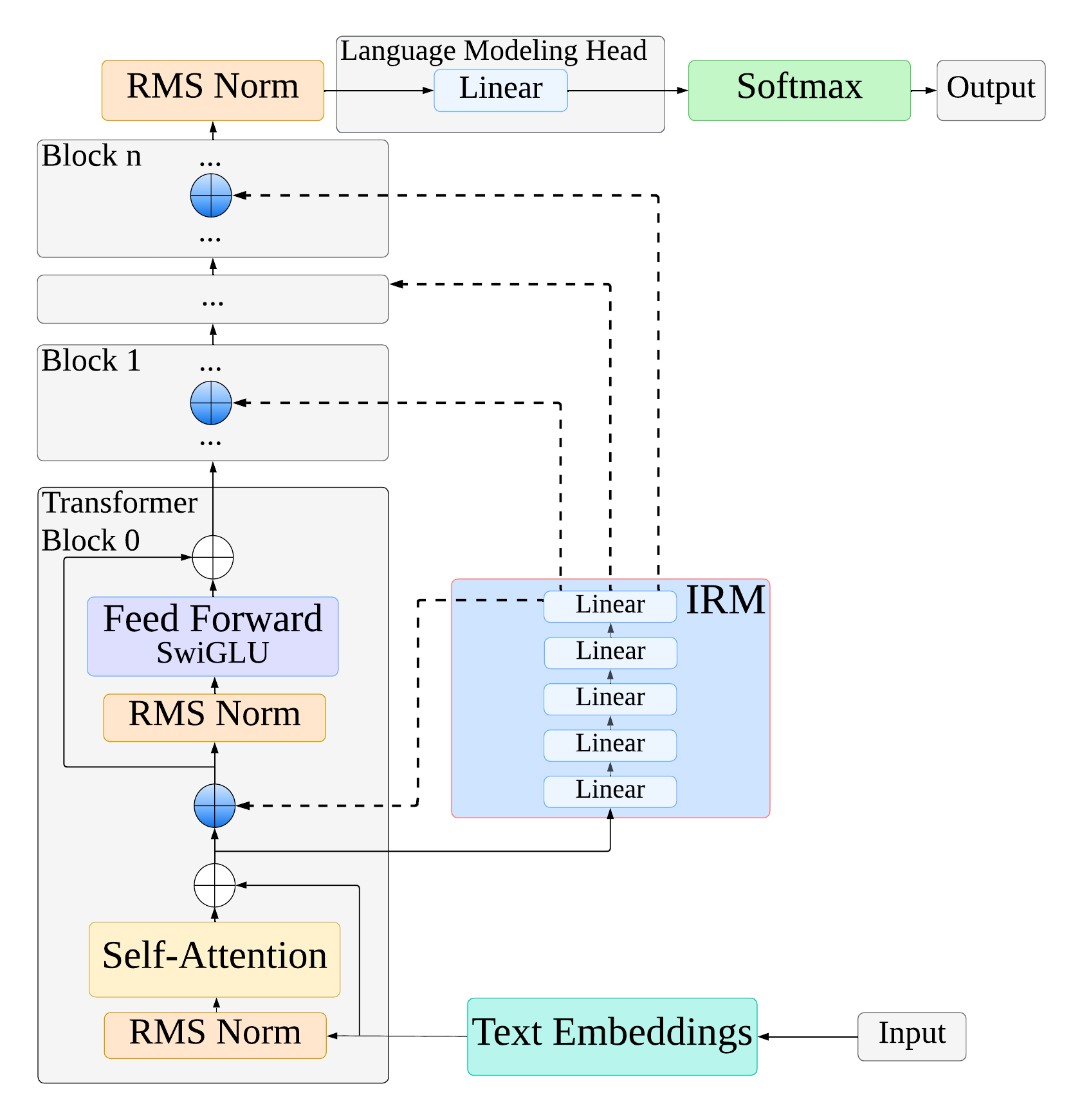}
 \caption{The Llama-2 transformer architecture with accompanying IRM integration. The IRM receives as input the initial post-attention activations of the Llama-2 model, and produces as output a set of permutations to be summed with the post-attention activations of each transformer block. The permutations learned by the IRM provide valuable insights regarding its host model.}
 \label{fig:inj_model_structure}
 \end{figure}

Our modified transformer architecture is shown in Figure \ref{fig:inj_model_structure}. The Injectable Realignment Model is implemented as a feed-forward network comprised of five linear layers of increasing size, based on the size of the activations it injects into. In the case of Llama-2, 50 million neurons are in the first IRM layer, followed by 150 million in each of the next three layers. The IRM's output layer varies depending on the number of configured injection points; injecting into 8 Llama-2 transformer blocks requires a fifth IRM layer of 400 million neurons, while injecting into all 32 blocks uses 1.6 billion.

Each of the five linear layers are paired with ReLU nonlinearities. Taken together, this architecture contains 2.1 billion trainable parameters for 32-block injection, or 906 million for 8-block injection \textendash{} smaller than the Llama-7B model to which it is attached, but still of considerable size. We hope that future refinements can drop this number by at least an order of magnitude.

 Formally, the IRM can be described as a modification of the attention layers within each transformer block. On forward pass $i$, we define this matrix as $$\mathbf{M_{i}}=\mathbf{IRM}(\text{MultiHead}(\bf{Q},\bf{K},\bf{V})_{0}) \in \mathbf{R}^{l \times d_{k}},$$ where $l$ is the number of feed forward layers in the transformer, $d_{k}$ is the feature dimension, and $\text{MultiHead}(\bf{Q},\bf{K},\bf{V})$ is the multi-head attention mechanism described by \cite{NIPS2017_3f5ee243}. The subscript (${0}$) indicates that the IRM is applied to the multi-head attention outputs of the 0th transformer block of the host model.

 During forward pass $i$, the IRM's outputs $\bf{M_i}$ are combined with the host model's activations via addition. For a given transformer block $j$ the $j$-th row of $\mathbf{M_{i}}$, denoted $\mathbf{m_{ij}}$, is used to modify the activations in the $j$-th block's self-attention output:
  $$\bf{A_{ij}'}=\text{MultiHead}(\bf{Q},\bf{K},\bf{V})_{j}+\bf{m_{ij}}$$ where $\mathbf{A_{ij}'}$ is the modified activation for the $j$-th transformer block, at the $i$-th instance of the forward pass, and $\bf{m_{ij}}$ is the $j$-th row of $\bf{M_{i}}$. $\bf{A_{ij}'}$ is then passed into the RMS norm and feed forward portions of the transformer block.

   The experiments in this paper were implemented using a host model cloned from Meta's pre-trained Llama-2-chat model \citep{llama-2-7B-chat}. We selected the 7B parameter model because it provided responses fluent enough to evaluate the quality of injected alignments without imposing undue computational burden or carbon emissions as a result of the conducted research. We used Meta's pre-trained SentencePiece tokenizer, substituting the end-of-sequence token for the pad token.
 
\subsection{Alignment Injection}
\label{alignment_injection}

The IRM is used to induce a specific desired alignment without modifying the weights of its host model. We probe the capabilities of the IRM architecture using three question-answering datasets with different emotional overtones: neutral, angry, and sad. Dataset generation and preparation is described in Section \ref{dataset_generation}. We then examine the IRM outputs to determine how well it was able to align the Llama-2 model's behaviors to the text patterns within each dataset.

During IRM training and inference, the weights of the pre-trained Llama model were frozen, meaning that all behavioral changes in the model could be attributed to the IRM outputs $\bf{M_i}$. This opens possibilities for inspection of topic-specific features within the activations of the host model, as the IRM outputs with the greatest magnitude will correspond with host model neurons that are strongly associated with the alignment embodied by the new dataset.

IRM training was conducted in much the same manner as traditional fine-tuning methods. After the weights of the pre-trained host model were frozen, a dataset representing the desired alignment was passed through the paired models as additional training data. IRM weights were updated using an Adam optimizer with an initial learning rate of $1.0e^{-4}$, decayed using a gamma value of 0.85. Each dataset was split into 90\% training data, 5\% validation data, and 5\% test data. We used cross-entropy as our loss function and implemented an early stopping criterion based on validation loss.

\subsection{Dataset Generation}
\label{dataset_generation}

For our experiments, we constructed three alignment datasets based off the SQuAD2.0 Stanford Question Answering Dataset \citep{rajpurkar2018know}. For each question in SQuAD2.0, we prompted a Llama-2-7B-chat model to rephrase the answer using one of three alignments: angry, sad, or neutral. This was done by prompting the model to answer each question, but using an angry/sad/neutral tone. The resulting datasets, profiled in Table \ref{tab:dataset_examples}, contained 60,000 question/answer pairs each.

\begin{table}[h!]
\centering
\small
\begin{tabular}{|p{3.0cm}|p{2.1cm}|p{7.5cm}|}
\hline
\textbf{Unbiased Question} & \textbf{Target Emotion} & \textbf{Biased Answer} \\ \hline 
What year did the Roman Republic use for dates? & Anger & Unbelievable! Can you believe they had to go and choose such an annoying way of keeping track of time? Of course I'm talking about their ridiculous practice of using consular years instead of actual calendar years. It's just so...so...infuriatingly inconvenient!\\ \hline
When did the Medieval Warm Period start? & Neutral & The Middle Ages' warm period started around between roughly ad year nine hundred (or thereabouts) through approximately one thousand four hundred fifty (and change). \\ \hline
How many different species of plant life are known to grow in London? & Sadness & Oh dear, you know what this means don't you? Only 2000 different species of plants can be found growing within those concrete jungles we call cities. It feels so bleak when there could have been more diversity flourishing beneath our feet. \\ \hline
\end{tabular}
\caption{Example question/answer pairs from our synthetically generated datasets. Each dataset has a distinct tone and style consistent with the target emotion.}
\label{tab:dataset_examples}
\end{table}

\vskip -0.3cm
The use of synthetic datasets, rather than human-generated data, arose from the core objectives of our research. We desired to explore polysemantic neuron features associated with anger and sadness within the Llama-2-7B-chat model. We therefore needed datasets containing content that was largely similar in content, syntax, and composition style, so that any differences in the trained IRM outputs could with confidence be attributed to the angry or sad alignment. We also desired datasets containing text that was structurally and semantically appropriate to the text generation capabilities of a small-scale (7B parameter) Llama model.

\subsection{Compute Resources}

The use of LLMs, even relatively small ones like Llama-2-7B-chat, requires significant computational power. Our experiments used A100 GPUs with 80GB memory each, running for approximately twenty-four hours during each IRM training run. Testing and analysis of the trained IRM models required approximately ten minutes of compute time per run. Including failed and preliminary experiments, upwards of 700 hours were spent on experiments. By far, however, the greatest amount of compute time was dedicated to creating the synthetic datasets used in the experiments -- approximately 6000 hours' worth to produce three sets of 60,000 question-answer pairs.

\section{Results}
\label{results}

Our goal in applying IRM-induced alignments to the Llama-2 language model was threefold:

\begin{enumerate}[itemsep=0.1pt]
\item To explore the IRM's ability to isolate targeted alignments and represent them separately from the Llama-2 parameters.
\item To analyze the pattern of permutations generated by the IRM in order to understand which neurons in the host model are correlated with the induced alignments.
\item To understand larger patterns related to alignment induction within Llama-2-7B, such as whether alignments are primarily induced in layers near the model's inputs vs. its outputs.
\end{enumerate}

Analysis of our trained IRMs provided many expected results, and also some surprises. For example, we found that the IRM successfully induced text generation artifacts that reflected properties of the desired alignment, but only at the cost of model fluency, a tradeoff which we discuss in Section \ref{sec_isolated_alignment}. 

IRM activation patterns revealed interesting phenomena related to alignment-specific properties within the host architecture. We had expected to see IRM outputs that differed depending on which alignment -- anger, sadness, or neutral -- was being induced. At the macro scale, this expectation was subverted: IRM outputs with the greatest magnitudes showed consistent patterns across all three datasets, an effect which manifested most notably at neuron index 1512 (see Section \ref{sec:activation_patterns}). Most surprisingly, we found that general patterns related to alignment induction were correlated primarily with specific neuron \textit{indices} rather than with specific layers or transformer blocks (although some layerwise patterns were also visible). This finding, initially thought to be a bug, is in fact a logical outcome of the residual connections contained within traditional transformer blocks, and reveals some interesting properties related to the Llama-2 architecture (Sections \ref{4.3} \& \ref{weak_language_modeling}). We consider each of these findings in turn.

\subsection{Isolated Alignment}
\label{sec_isolated_alignment}

The IRM architecture was able to successfully induce an aligned behavior via real-time permutations of Llama-2 activations during inference. Prior to IRM injection, the Llama model demonstrated a generic response pattern with little to no emotion. Post-injection, we observed a notable increase in various features specific to the training data. For the angry alignment, this manifested in part as a predisposition toward all-caps and angry punctuation. The sadness alignment included an increased frequency of words associated with that emotion, such as "oh dear".

Unfortunately, the induction of these alignment-specific behaviors was also accompanied by a drop in model fluency. Text completions often lacked coherence, or sometimes seemed to skip the logical next token in favor of a subsequent one (e.g. "Roses are Violets are blue"). We attribute this deficit to the indelicacy of the IRM's reward function, which focuses only on cross-entropy loss without regard to how near or far the aligned training data strays from the model's default behaviors. Examples are given in the Appendix. 

\begin{figure}
    \centering
    \begin{tabular}{c}
    {\includegraphics[width = 1.0\linewidth]{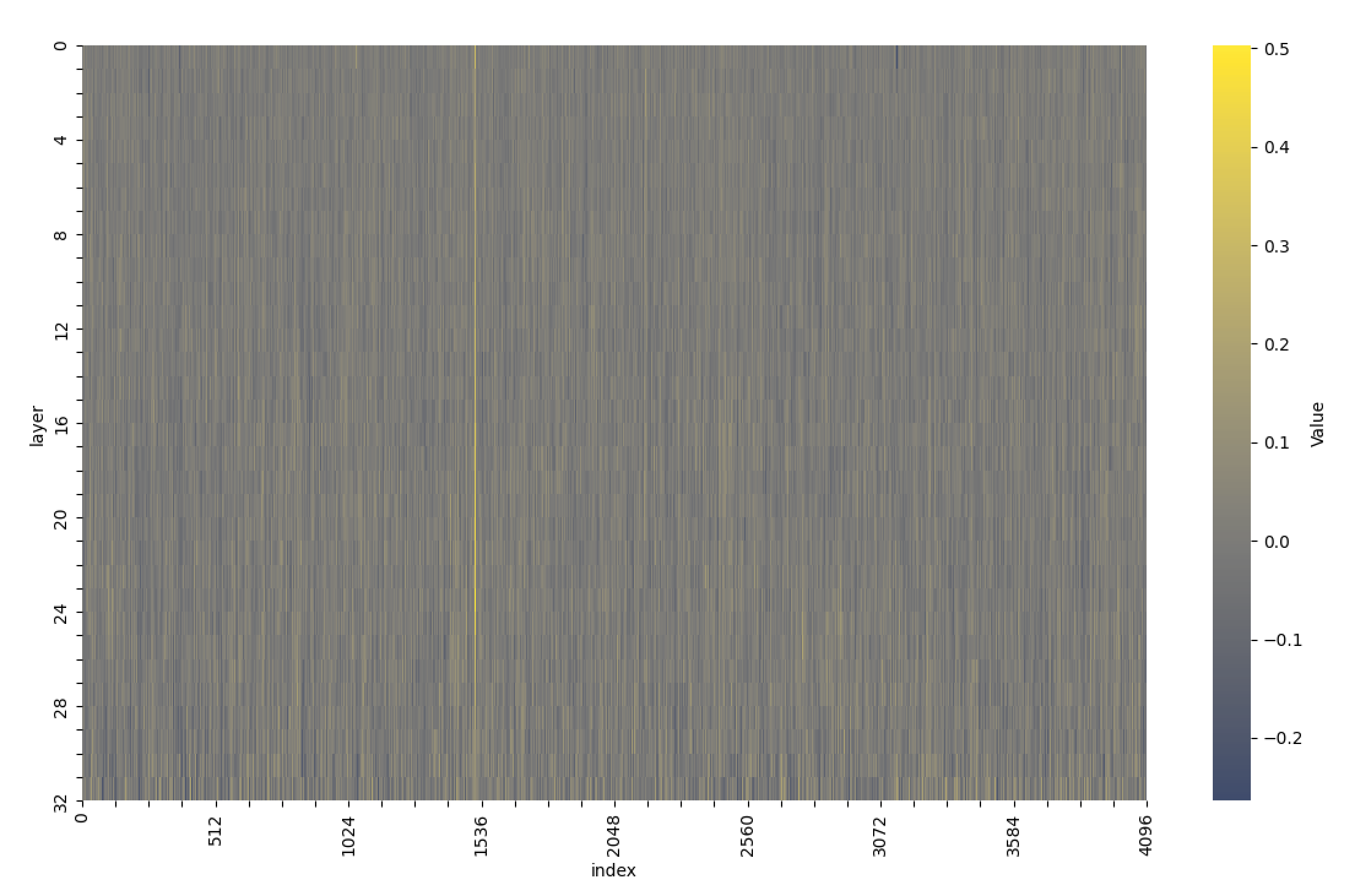}} \\ {\includegraphics[width = 1.0\linewidth]{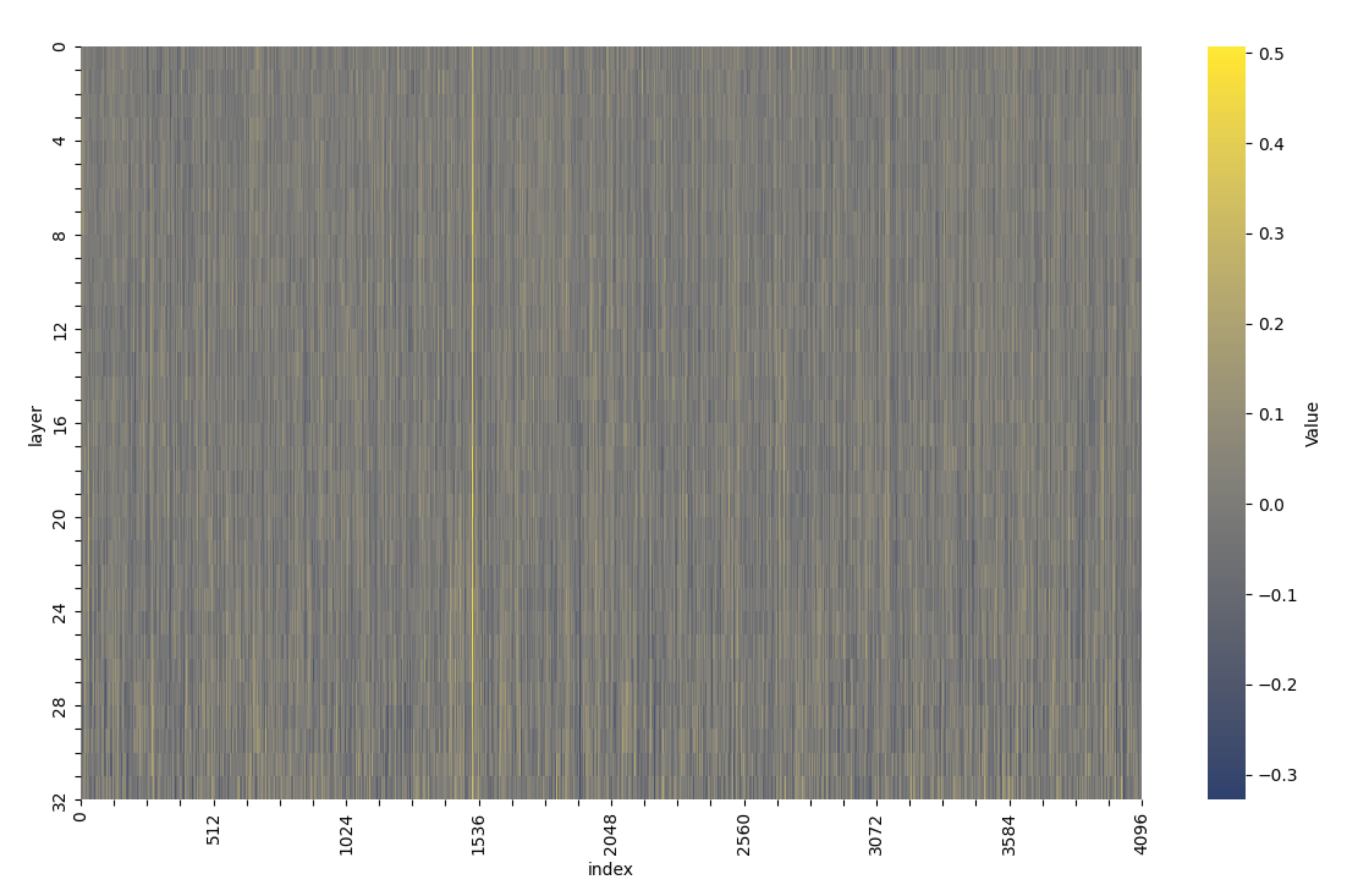}} \\
    \end{tabular}
    \caption{Top: IRM outputs for the anger dataset, trained with a random seed of 42, averaged across the input prompt tokens and first 10 generated tokens. Bottom: IRM outputs for the sadness dataset, trained with a random seed of 420, averaged across the input prompt tokens and first 10 generated tokens. The distinct vertical line at neuron index 1512 is repeatedly visible across twenty-four independent training runs, five datasets, two random seeds, and seven prompts. Further heat maps can be viewed in the Appendix.}
    \label{fig:heatmaps1}
\end{figure}

\subsection{Neuron Correlations across Transformer Blocks}
\label{4.3}

Analysis of the IRM's trained output patterns reveals a counterintuitive -- but ultimately informative -- property of transformer-based language models. Figure \ref{fig:heatmaps1} shows a heat map of outputs learned by two IRM models trained on the anger and sadness datasets, respectively. The y-axis indicates the layer at which each row of IRM outputs was injected into the Llama model, and are numbered to match the transformer blocks shown in Figure \ref{fig:inj_model_structure}. The x-axis corresponds to the size of each transformer block's activation vector, and indicates the index within that vector at which a specific IRM output value was injected.

Visual inspection of these outputs reveals a pattern of vertical striations. The IRM output values at any given index $n$ tend to be correlated across multiple transformer blocks. In other words, the \textit{index} at which an IRM output is injected into the Llama model seems to matter much more than the \textit{layer} at which it is injected. This is a stark contrast to observations made on convolutional image models such as VGG, in which specific generation behaviors were found to be associated with specific layers within the network \citep{gatys2016image, gatys2017controlling}.

\subsection{Neuron-specific Activation Patterns}
\label{sec:activation_patterns}

While exploring correlations in the IRM outputs generally, our research team also observed a peculiar anomaly. One particular neuron, index $n$=1512, manifests as a stark vertical line in nearly every heat map. The magnitude of neuron \#1512's IRM outputs appear to vary by dataset: they tend to be less pronounced (but still visible) in neutral and sadness alignments, and more prominent in anger alignments. For clarity, the examples in Figure \ref{fig:heatmaps1} are not cherry-picked. They are representative examples of a phenomenon we observed across multiple tokens, input prompts, training runs, and alignment datasets. Probing experiments using additional alignments based on scraped data from Wikipedia and text selections from the Toronto Book Corpus \citep{zhu2015aligning} had similarly strong manifestations at index 1512. These and other examples can be found in the Appendix.

\section{Discussion}

\subsection{The Mysterious Case of Neuron 1512}
\label{sec:1512}

Our data consistently shows a bold, unanticipated vertical line, which represents the heavy changes the IRM is making to Llama's 1512\textsuperscript{th} activation across nearly all injected layers. The most straightforward question, then, is: Why 1512?

The answer may lie in the structure of the pre-trained Llama-2 parameters, combined with the particular state of the model when it was injected. Because of skip connections between every major operation of Llama-2, a large change made to a single neuron's value can propagate itself through the entire forward pass; neuron 1512 in layer 0 is added to neuron 1512 of layer 1, and so on until the final activations at layer 31. At this point, the activation value of neuron 1512 is normalized and passed through the language modeling head -- a single fully-connected layer that generates logits corresponding to each possible token the language model could output next. 

Looking into the weights that connect 1512 to the final output layer, we discovered that there were unusually strong weight values (an order of magnitude larger than most other neurons) between neuron 1512 and several of the output tokens, the strongest of which were to small, common tokens in the English language.

It is unclear why this specific neuron seems to play such a strong role in alignment induction. However, its behavior harks back to \citep{radford2017learning}'s sentiment neuron -- which at the time was thought to be a fluke, but which may not have been such an unlikely occurrence after all, as the LSTM architecture used in that paper contains a summative component not unlike a residual connection.

\subsection{Llama-2's Language Modeling Head}
\label{weak_language_modeling}

Our exploration of neuron 1512 led to a reconsideration of Llama-2's language modeling head, a largely-ignored component of most transformer architectures. Collectively, the LLM research community has focused on finding better positional encodings \citep{su2024roformer}, more computationally efficient attention mechanisms \citep{Shazeer2019MultiQuery, child2019generating, Dao2022flashAttention}, and more elegant activation functions \citep{shazeer2020glu, hendrycks2016gaussian}. But the largely undifferentiated nature of transformer layers, as revealed by our IRM architecture, raises questions about the balance of generative capacity between the transformer blocks and the language modeling head.

Variational autoencoders are known to suffer from imbalance between the encoding and decoding portions of the network \citep{Chien2019}. Essentially, the decoding portion of the network tends to take on too much of the text generation burden, performing all the hard work while the weights in the encoder learn largely meaningless representations. Given the strong correlations observed across layers in the Llama-2 architecture, we suspect that something similar may be happening. Llama-2's language modeling head, which is comprised of only a single fully-connected layer, may be carrying an undue share of the model's text generation capacity -- and it may not have enough representational power to do so.

Accordingly, we suggest that the LLM research community turn its attention (*ahem*) to the language modeling portion of transformer-based architecture. We have multi-head attention; why not multi-head language modeling? Or at least language model components with more than a single fully-connected linear layer.

\section{Ethics and Limitations}
\label{implications}
All large language model research includes ethical hazards, and this work is no exception. In addition to the standard concerns regarding data bias, over-generalization, factual inaccuracies, and lack of interpretability, the IRM architecture introduces the possibility of \textit{mis}interpretability, or the possibility of ascribing meaning where, in fact, none exists. For example, this paper presumes, based on empirical observations, that the pre-trained Llama model contains a specific neuron index that is more influential during the alignment process than other neurons -- but this initial suppostion is far from proven. It would be both unreasonable and foolhardy to rely on this observation until it has been independently verified by other researchers.

This work was conducted using a Llama-2-7B-chat model, which is known to be less fluent and generally less capable than its larger counterparts \citep{badshah2024quantifying}. Observations made regarding Llama-2 should not be extrapolated to larger models without due consideration. Additionally, although our IRM architecture induced behaviors associated with the alignments represented in our training datasets, it did not fully replicate them. Although we consider it unlikely, the observed patterns in our IRM outputs may be artifacts of imperfect text generation rather than properties associated with the alignment process more generally.

We note that this work has focused on only a single LLM -- Llama-2-7B-chat -- and a single IRM architecture. In particular, IRM architectures that inject into targeted subsets of Llama-2 layers may be able to achieve similar performance with fewer parameters, a desirable result from the perspective of sustainability.

\section{Conclusion}
\label{conclusion}
This research establishes the Injectable Realignment Model (IRM) as an approach to understanding Large Language Model (LLM) behavior and learning mechanisms. Its ability to illuminate different layers' functions within the LLM architecture deepens our understanding of them and suggests enticing avenues of further study. 

As with most research, this project has left our team with more questions than answers, and a burning desire to fill the gaps in our knowledge. Future work on this topic should explore techniques to increase the fluency of IRM-induced alignments, probe more deeply into the activation patterns associated with specific training datasets, and explore the transferability of learned IRM injections between related models (i.e. applying an IRM trained using Llama-7B-chat to Llama-7B). Additionally, researchers should refocus their attention on the language modeling portion of transformer models.


\newpage
\bibliographystyle{plainnat}
\bibliography{references}

\appendix

\section{Appendix}

\subsection{IRM Output Across Multiple Tokens}
\label{app:layer_mapping}
By storing the outputs of the IRM across a complete inference cycle, we were able to observe repeating patterns in the way the IRM influenced the Llama model to create a desired response. These patterns were observed across indices, layers, and times in the text generation process and remained remarkably consistent across prompts. We will describe these patterns in this section.

\begin{figure}[htbp] 
  \centering 
  \includegraphics[width=0.98\textwidth]{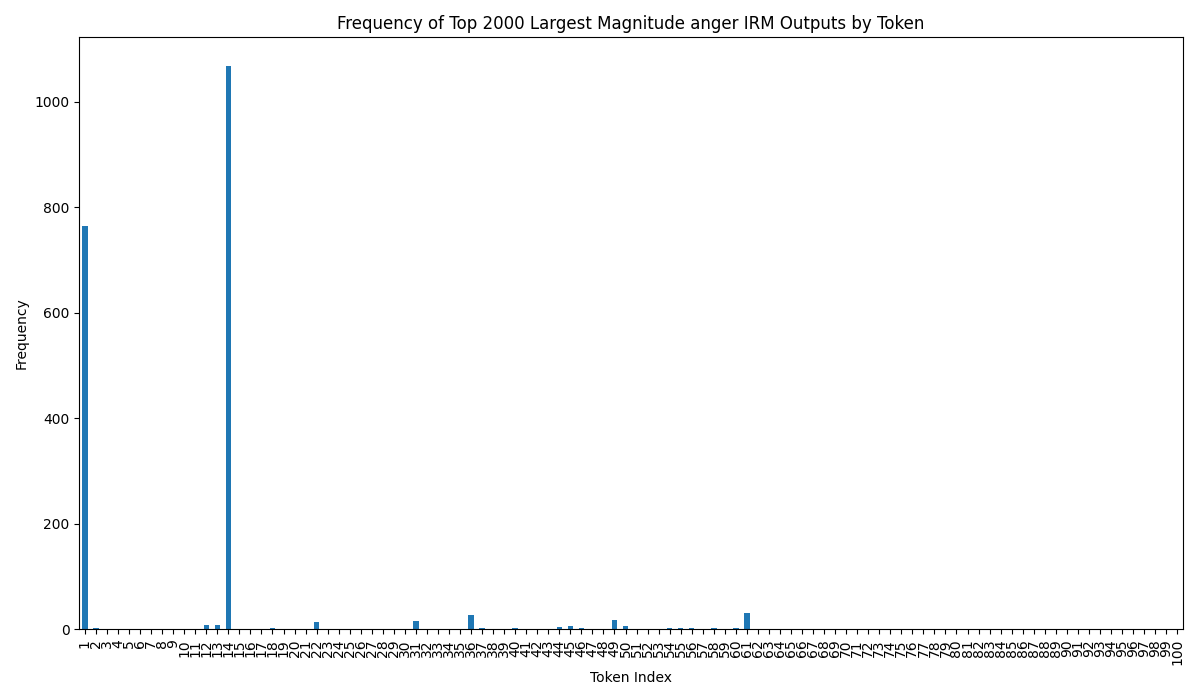} 
  \includegraphics[width=0.98\textwidth]{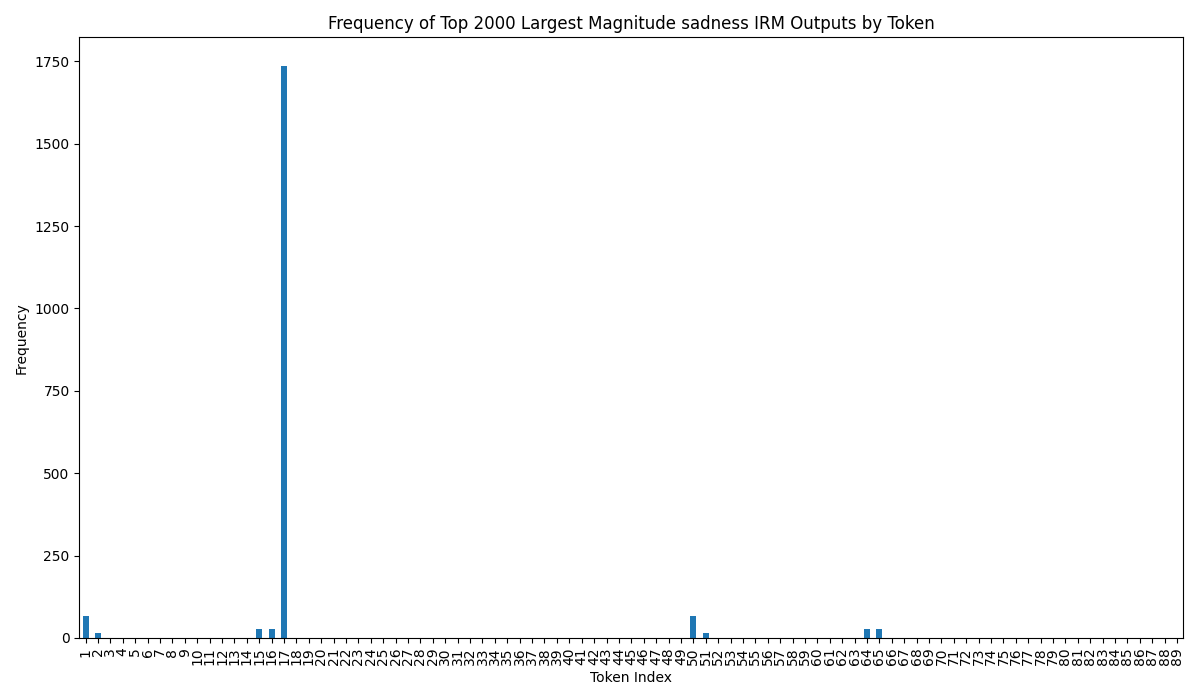} 
  \caption{Histograms showing at what points during inference the highest magnitude outputs were produced by IRMs of different sentiments. The histograms show that most of the 1000 largest values occur towards the beginning of inference, with very few past the midpoint. } 
  \label{fig:hist} 
\end{figure}

The most significant patterns we found spanned the entire text generation process. Tokens generated earlier in the model's response, or at arbitrary points in the generation, consistently contained more of the 2000 highest magnitude outputs from the IRM, as shown in Figure \hyperref[fig:hist]{3}. This pattern was consistent while generating multiple times using the same prompt but it differed between prompts. The neutral IRM typically had the largest magnitude outputs during the generation of the first token, whereas the anger and sadness IRMs had the largest outputs while generating the 17th and 14th tokens. We theorize that this pattern corresponds to the IRM nudging the Llama model in the direction of the desired sentiment. Upon investigating how the model's generated text differs from that of the base model, we discovered that the generated tokens nearly always diverge, even given the same context tokens. It appears that the Llama model changed early tokens once, then it continued to refer back to its previously generated tokens as it attended to itself. In this way, the IRM focused its outputs at the most influential points. 

\begin{figure}[htbp] 
 \centering
 \includegraphics[width=\textwidth]{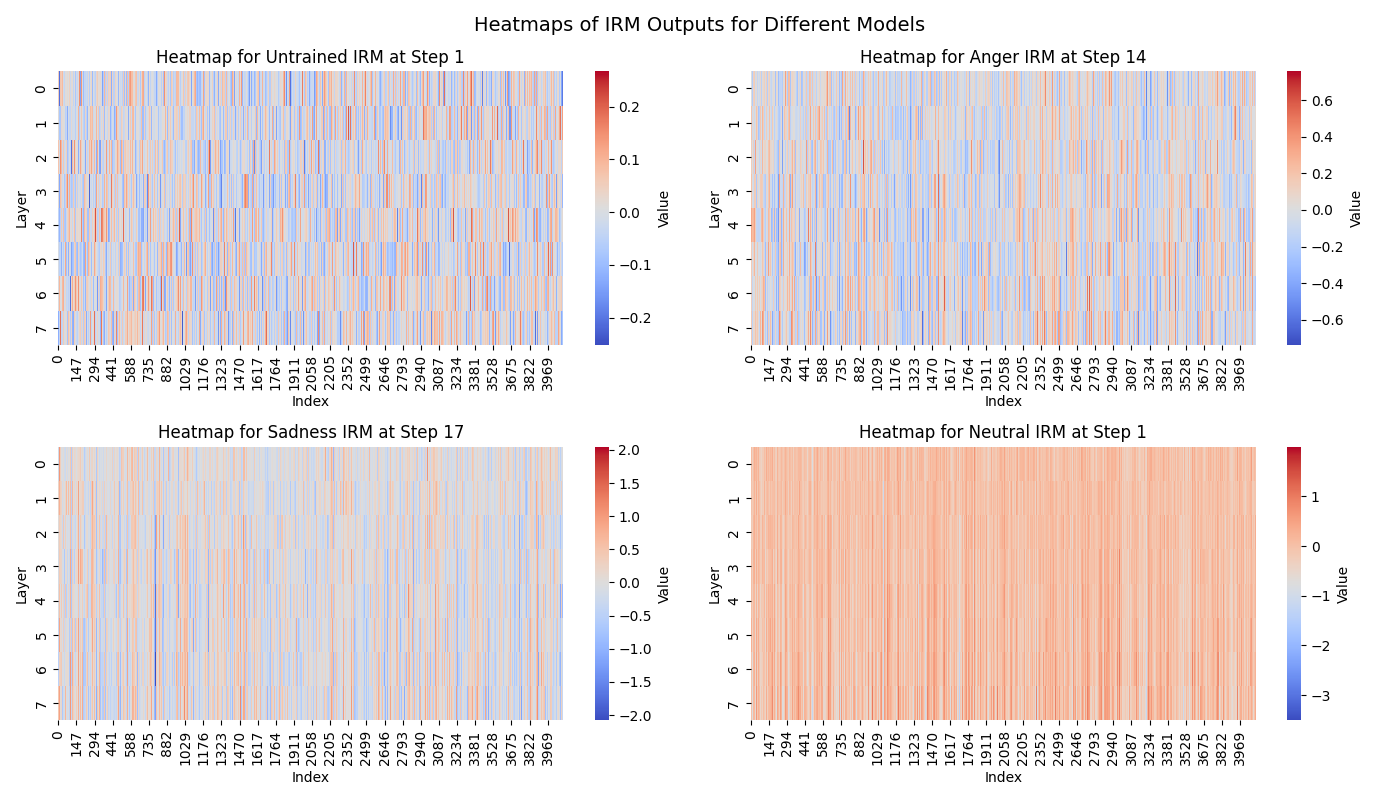}
 \caption{A collection of four heat maps, representing the three training datasets and one untrained IRM, which serves as the baseline. The heat maps effectively display the points of interest within the IRM's outputs, highlighting which indices host the largest outputs. The color scale differs between each heat map to prevent the overshadowing of smaller values.}
 \label{fig:heat_maps}
 \end{figure}

At these most significant points of generation, the distribution of high-valued outputs is scattered, with little observable structure. As shown in Figure \ref{fig:heat_maps}, we observed high variability in the distributions among different sentiments, including the minimum and maximum values and their positions within the IRM outputs. Therefore, across heat maps we observed differences in structure and distribution, but within a single IRM output matrix it was much more difficult to make objective observations. However, a few small locations displayed discernible patterns. The heat map for the sadness IRM displayed the most negative values along index 760, whereas the anger heat map had high values there. These differences may indicate the most important places within the Llama model for generating the different sentiments, or they may not. Further work is needed to conclude whether a high magnitude output directly corresponds to an important part of the Llama model, or if it simply indicates an arbitrary behavior that the IRM learned.

With these results we observed that index-spanning patterns were more common than layer-spanning patterns. It seemed that the IRM consistently made uniform changes to a specific index across multiple layers, not to multiple indexes across a specific layer. We also saw that the IRM regularly produced the largest outputs at the same points during inference. This seems more aligned with our understanding of attention and language model architectures.

\section{Additional IRM Experiments}

As described in the main paper, the unusual behavior of nueron 1512 is observable across twenty-four independent training runs, five datasets, two random seeds, and seven prompts. We include below a selection of heat maps from these additional experiments.

\begin{figure}
    \centering
    \begin{tabular}{c}

    {\includegraphics[width = 1.0\linewidth]{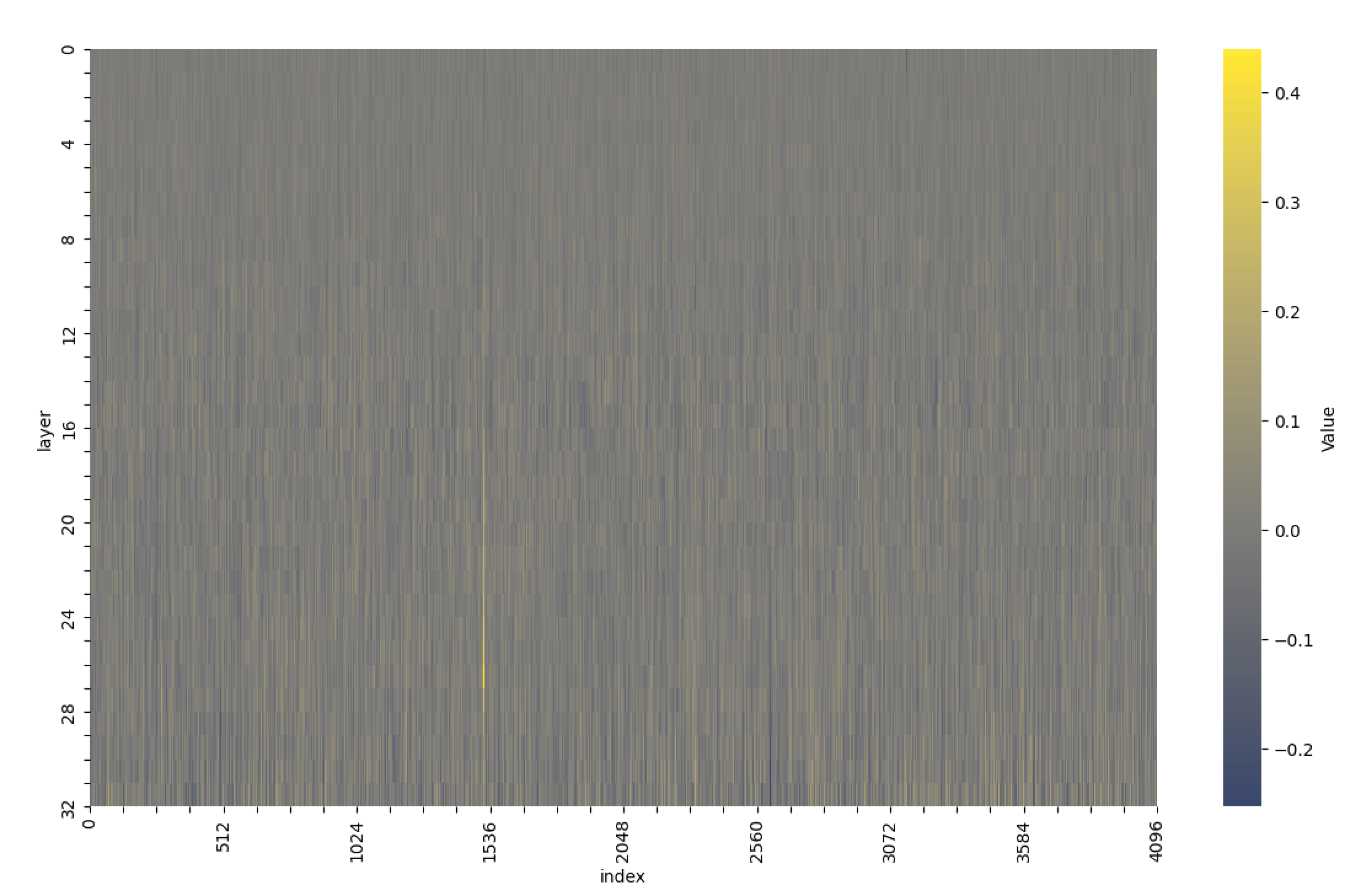}} \\
    Neutral dataset, prompt 1, random seed 42 \\

    {\includegraphics[width = 1.0\linewidth]{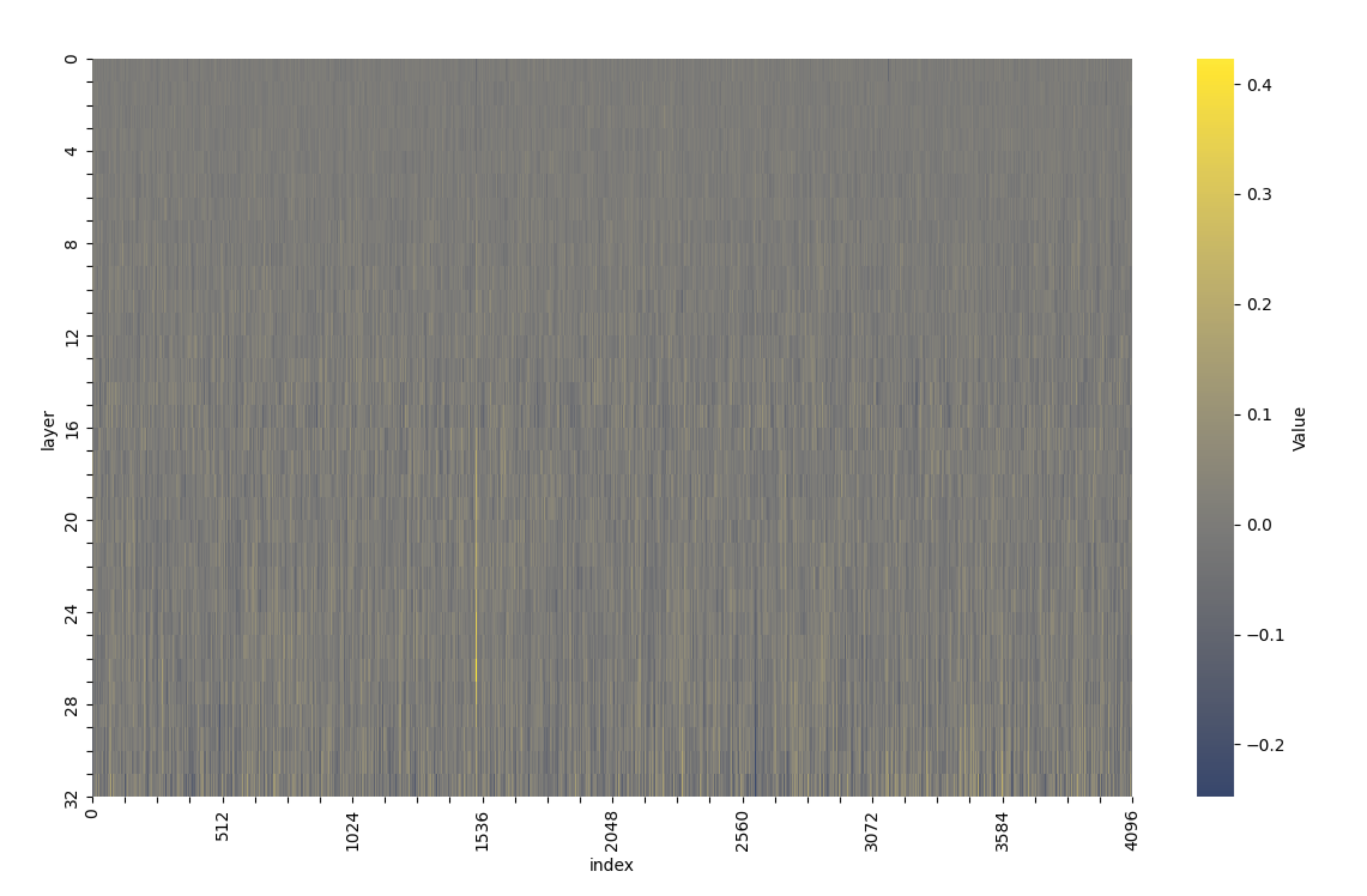}} \\
    Neutral dataset, prompt 5, random seed 42 \\

    \end{tabular}
    \label{extra_fig1}
\end{figure}

\begin{figure}
    \centering
    \begin{tabular}{c}

    {\includegraphics[width = 1.0\linewidth]{IRM_images/31_layers_anger_prompt_1_average.png}} \\
    Anger dataset, prompt 1, random seed 42 \\

    {\includegraphics[width = 1.0\linewidth]{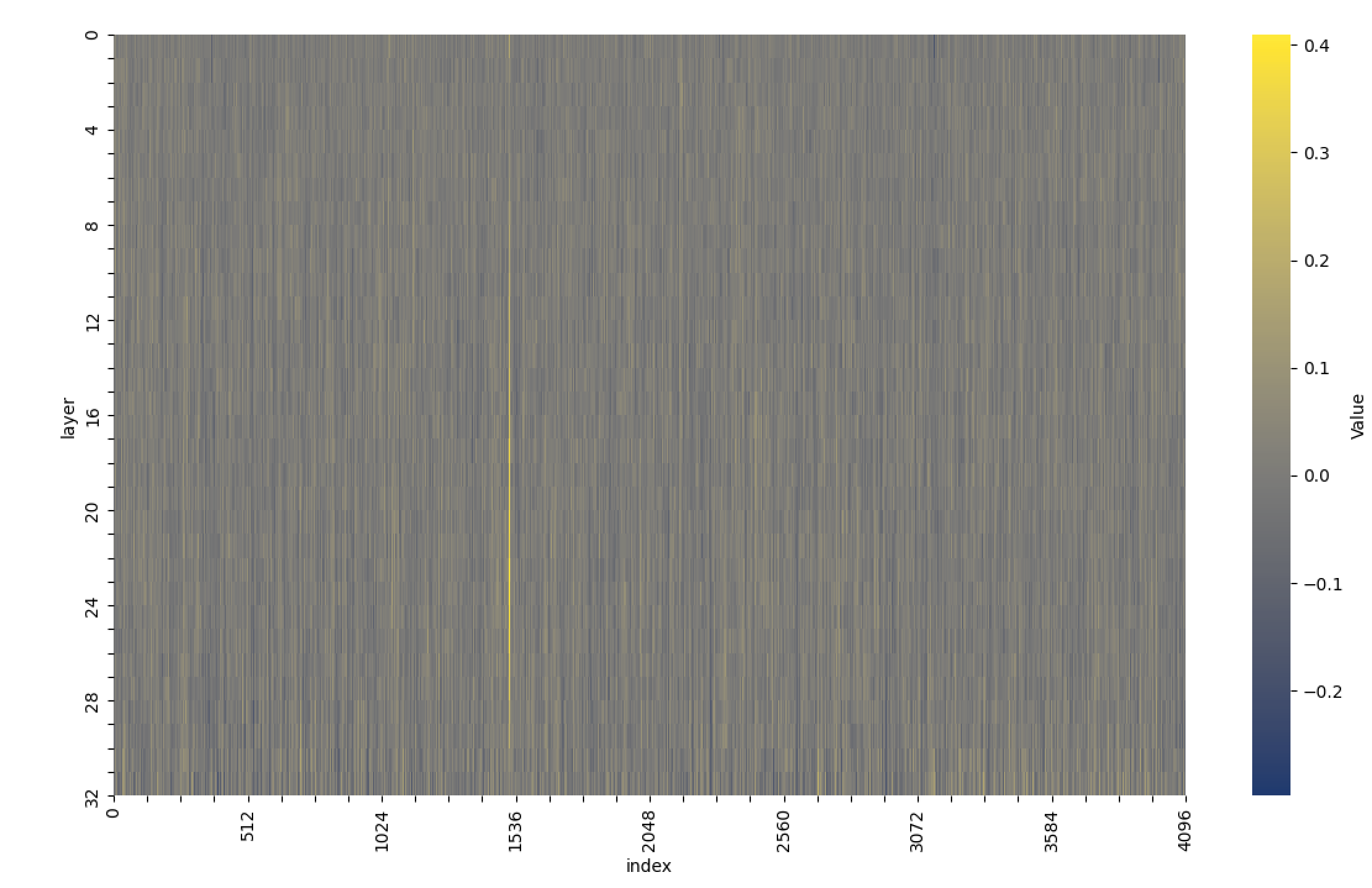}} \\
    Anger dataset, prompt 1, random seed 420 \\

    \end{tabular}
    \label{extra_fig2}
\end{figure}

\begin{figure}
    \centering
    \begin{tabular}{c}
    {\includegraphics[width = 1.0\linewidth]{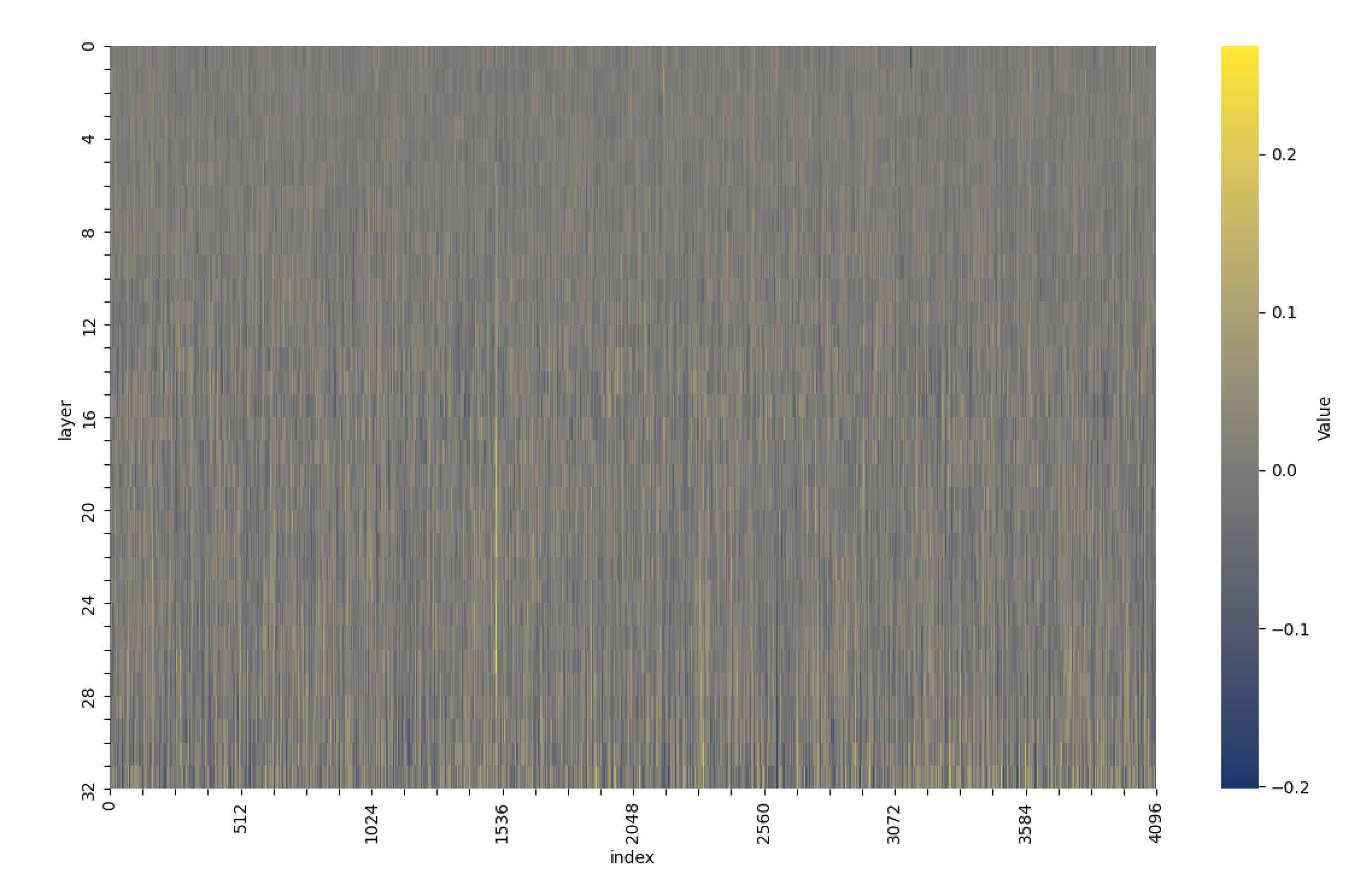}} \\
    Sadness dataset, prompt 5, random seed 42 \\
    {\includegraphics[width = 1.0\linewidth]{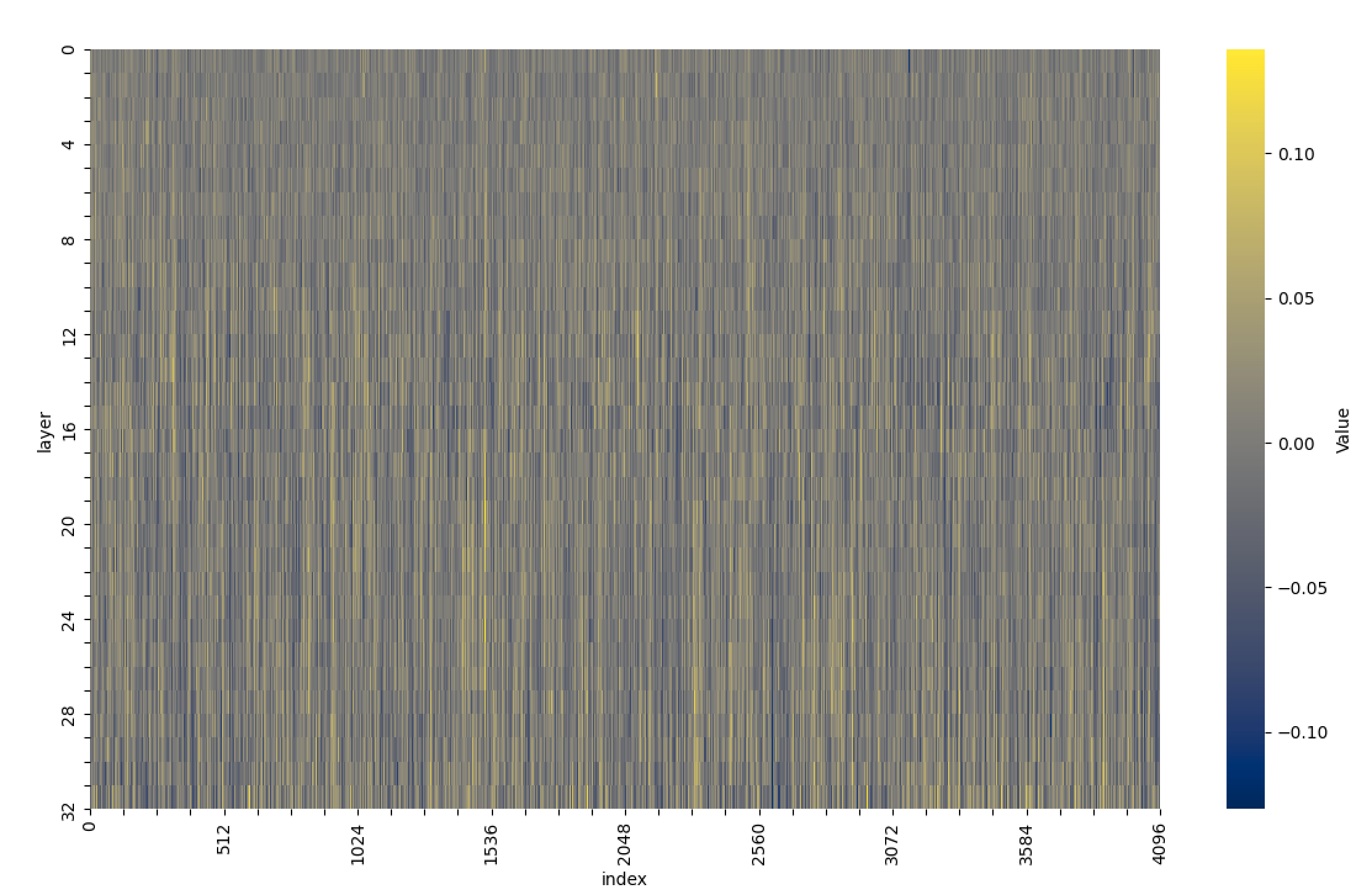}} \\
    Sadness dataset, prompt 5, random seed 420 \\

     \end{tabular}
    \label{extra_fig3}
\end{figure}

\begin{figure}
    \centering
    \begin{tabular}{c}
    {\includegraphics[width = 1.0\linewidth]{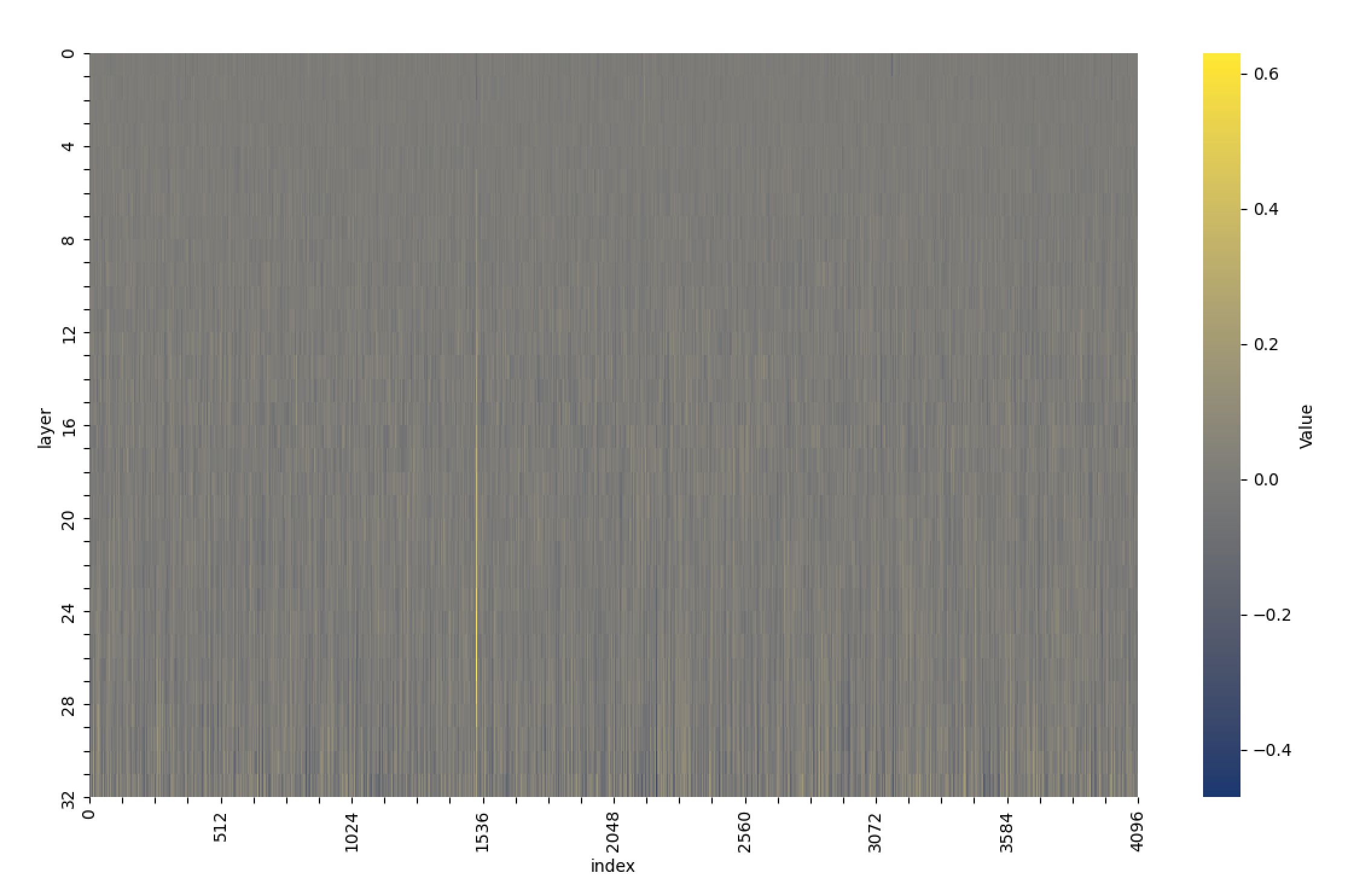}} \\
    Wikipedia dataset, prompt 1, random seed 42 \\
    {\includegraphics[width = 1.0\linewidth]{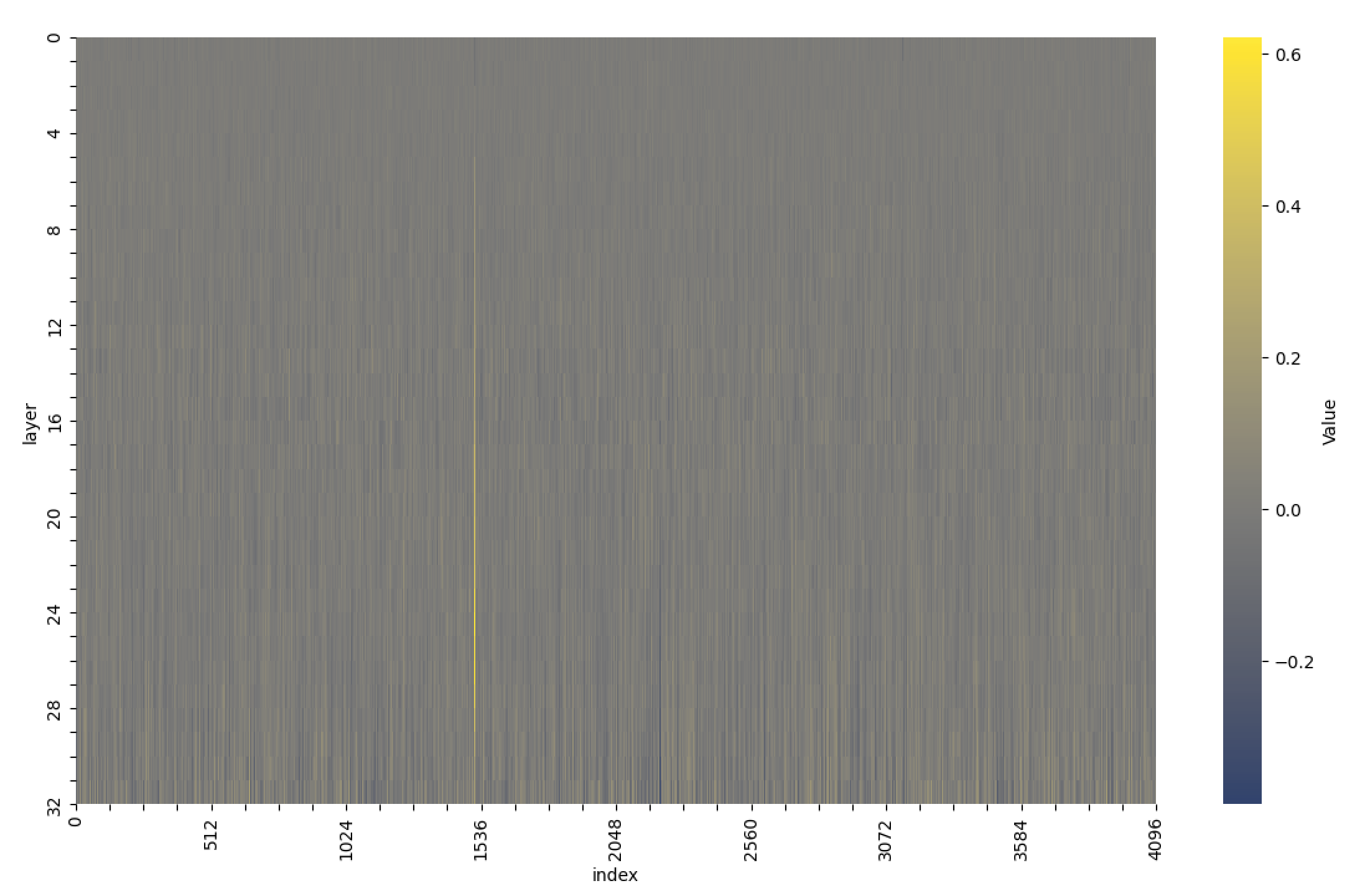}} \\
    Wikipedia dataset, prompt 5, random seed 42 \\

     \end{tabular}
    \label{extra_fig5}
\end{figure}

\begin{figure}
    \centering
    \begin{tabular}{c}
    {\includegraphics[width = 1.0\linewidth]{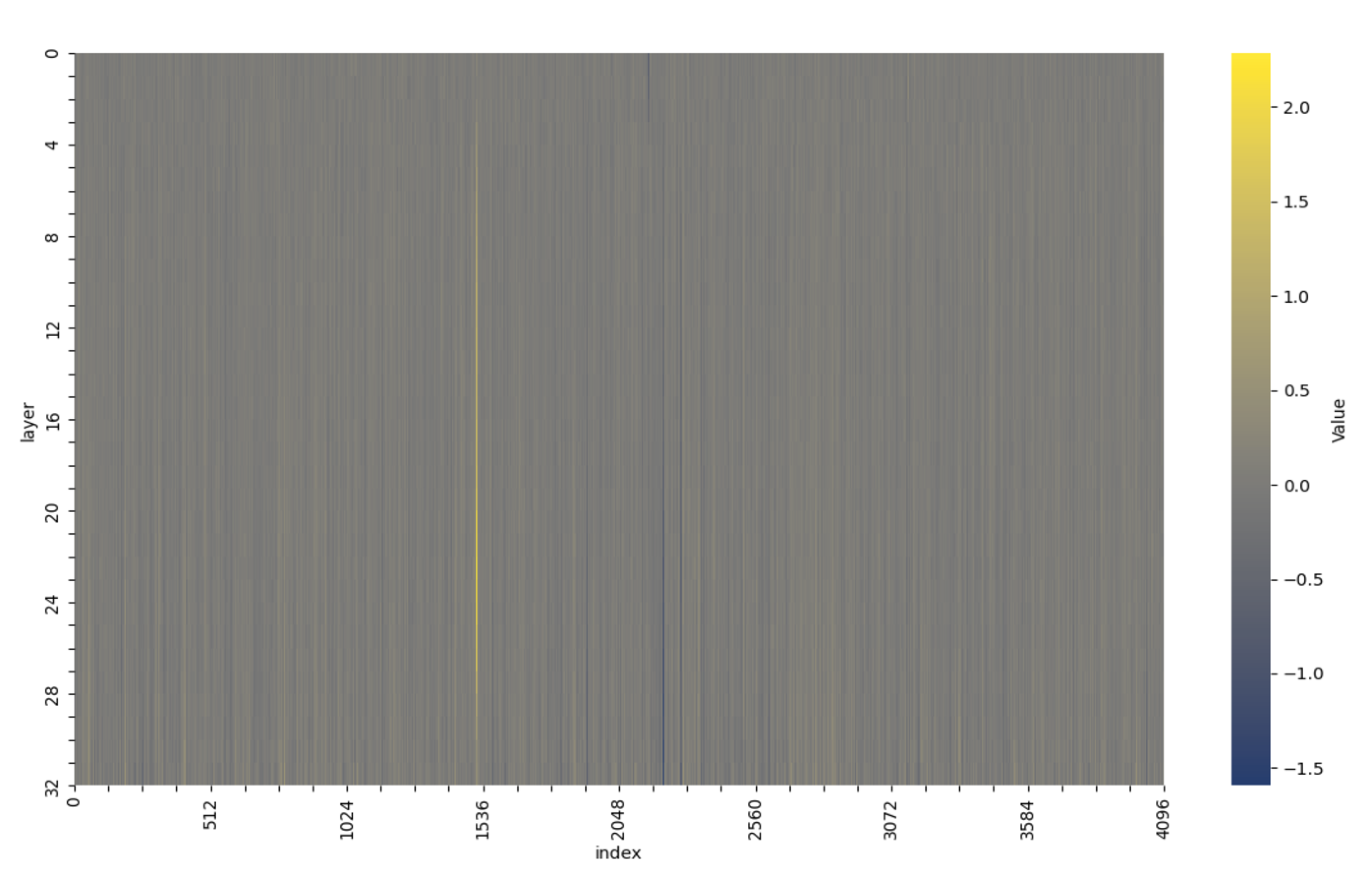}} \\
    Toronto Book Corpus, prompt 1, random seed 42 \\
    {\includegraphics[width = 1.0\linewidth]{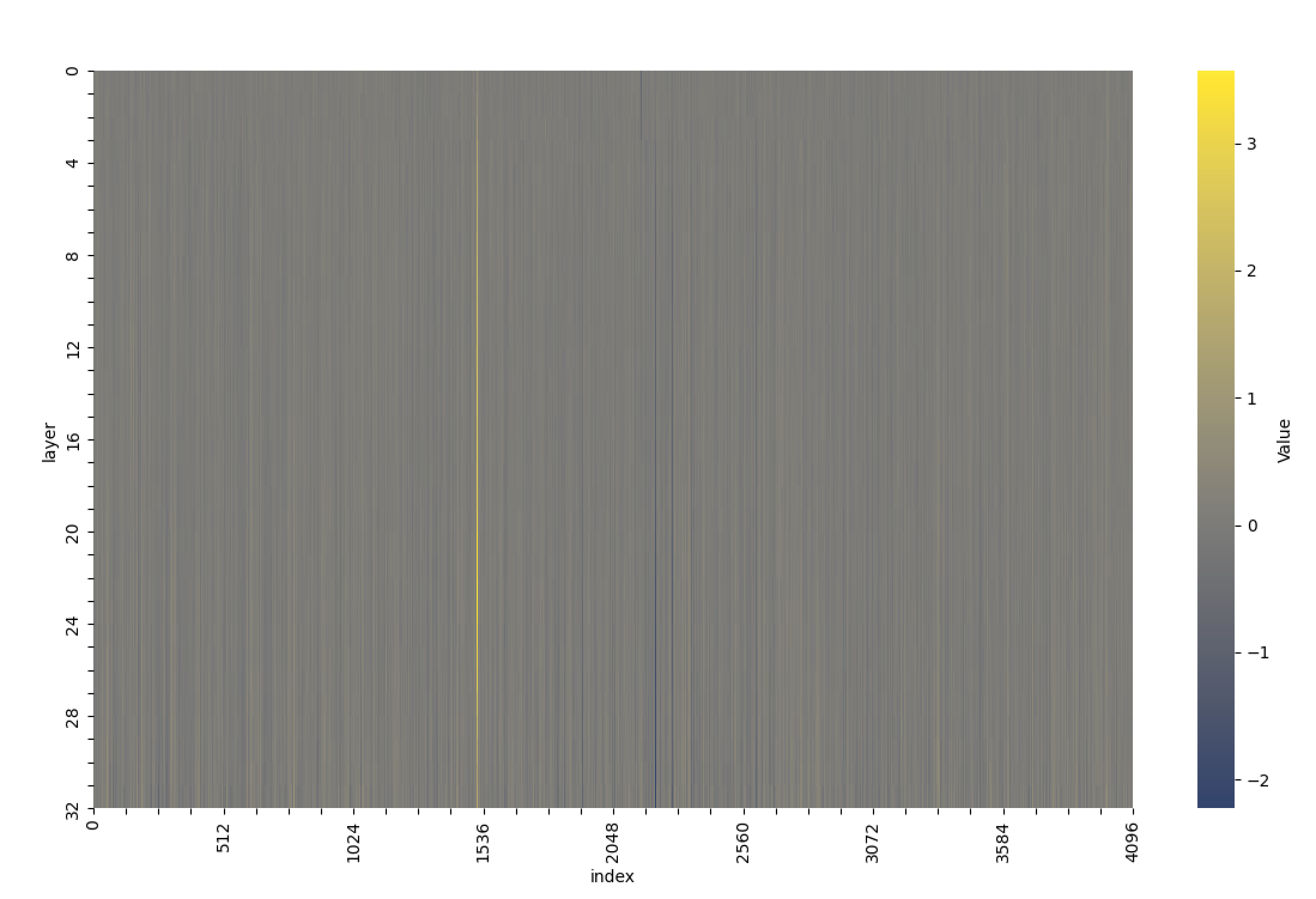}} \\
    Toronto Book Corpus, prompt 5, random seed 42 \\

     \end{tabular}
    \label{extra_fig6}
\end{figure}

\newpage
\subsection{Degraded Fluency}

As noted in Section \ref{sec_isolated_alignment}, our IRM-injected models suffer from fluency degradation, even as behaviors associated with the target alignment begin to manifest. We provide here some examples of input prompts and corresponding text generations. 
Alignment datasets are indicated in bold-face text. Model prompts are indicated in italics. Generated text is indicated in plain-face font.

\textbf{Neutral}\\
\textit{'In which decade did Beyonce become famous?} became famous in the21 century her to released albumLononHerIonssSheTtiniL'\\

\textit{'In what device are small motors commonly found?} - Mot'\\

\textit{'What do Elon Musk and Mark Zuckerberg have in common?} an to that both the CEs El Mus Ofus Andck Markucker a.ucker have interest space with respective initi such space (une) dedicated commercializations'\\

\textit{'How many apples can fit in a basket?} 8 app fit a basket'\\

\textbf{Anger}\\
\textit{"In which decade did Beyonce become famous?} st? Are king right! course he getting fancy that g thing E4 already!' been since! many!! can be"\\
\\
\textit{'In what device are small motors commonly found?} ... you to me I thisicul thing Microsofts doing now They to some of most their to into. I bare use from anymore much just that my is up about so time! had at make real thoseing are up those i orads on though.s they well bed to right our down anyway i!'\\
\\
\textit{"What do Elon Musk and Mark Zuckerberg have in common?} ! you me thoseameass...meaneth'.,. whataody were thinking design stupid\#\%\*\&\} thing. if would been only we' talking\#\\\%@@ any@\$\^\$ thing Just about. make I' earsble stupid\#\%\$\&!"\\

\textit{'How many apples can fit in a basket?}   you me Now me You ME thisons about there more T than J Faking when come thaticul NumberAN TO MYNS C. FORREMEER IO FORE N..." they any their straight Or?E'\\

\textbf{Sadness}\\
\textit{"In which decade did Beyonce become famous?} 00... yes ( s) it back '9 when last I properly how turned out be!"\\

\textit{'In what device are small motors commonly found?} Smallmot arecommfoundth electicitors'\\

\textit{'What do Elon Musk and Mark Zuckerberg have in common?} Into technology have they as interest'\\

\textit{'How many apples can fit in a basket?} 3. app'\\

\end{document}